\documentclass{article} 
\usepackage[final]{colm2026_conference}

\usepackage{microtype}
\usepackage{hyperref}
\usepackage{url}
\usepackage{booktabs}
\usepackage{amsmath,amssymb,amsfonts}
\usepackage{amsthm}

\usepackage{graphicx}
\usepackage{textcomp}
\usepackage{hyperref}
\usepackage{url}
\usepackage{booktabs}
\usepackage{amsfonts}
\usepackage{nicefrac}
\usepackage{microtype}
\usepackage{xcolor}
\usepackage{adjustbox}
\usepackage{algorithm}
\usepackage{algpseudocode}

\usepackage{enumitem}
\usepackage{tabularx, caption}
\usepackage{MnSymbol}
\usepackage{subcaption}

\usepackage{lineno}

\definecolor{darkblue}{rgb}{0, 0, 0.5}
\hypersetup{colorlinks=true, citecolor=darkblue, linkcolor=darkblue, urlcolor=darkblue}

\title{PB-GRPO: Learning Socially Adaptive LLM Agents from Persona-Driven Simulation
with Preference-Batched GRPO}

\author{
Jim Wang\thanks{Work completed during a 2025 internship; correspondence: Bin Guo, \texttt{guobg@amazon.com}.},
Jun Yin, Xu Han, Yongsheng Mei, Jie Hao, Bin Guo \\
\small Amazon
}

\begin{document}

\ifcolmsubmission
\linenumbers
\fi

\maketitle

\begin{abstract}
Building LLMs that behave well \emph{socially}, not merely correctly, requires more than producing locally helpful responses. A socially competent agent must infer users' unstated goals, respect their preferences, and adapt as the conversation unfolds.
These behaviors are inherently \emph{multi-turn} and \emph{social}, making them hard to optimize: real interaction data is scarce, and user preferences are typically \emph{latent} rather than directly observable. To address these challenges, we build on a persona-driven social simulation environment (consisting of a persona library, LLM-based user simulators, and a user-satisfaction scoring system ranging from $[0,1]$), to introduce preference-batched GRPO (PB-GRPO), a post-training algorithm that learns socially adaptive policies from conversation-level feedback. Compared to vanilla GRPO, PB-GRPO computes advantages using a normalization estimated across a bucket of users with similar preferences, stabilizing training across a diverse social population. Empirical evidence shows that PB-GRPO improves models' social behavior over strong reinforcement-learning baselines in our simulated environment.
\end{abstract}

\section{Introduction}
LLMs are increasingly deployed as conversational partners, such as assistants and role-playing agents. In these settings, users judge them not only on \emph{correctness} but on \emph{social competence}: the ability to behave well in an interaction with another person. A socially competent agent must infer what users actually want, respect their constraints, adapt its strategy as the user reacts, and decide when to ask a clarifying question rather than act immediately. 
Social behavior is hard to learn because user preferences are often \emph{latent}, not stated outright \citep{shani2024multiturnreinforcementlearningpreference}. Prior work distinguishes \textbf{explicit} social cues (directly stated likes/dislikes or facts) from \textbf{implicit} ones (inferred from behavior and conversational style) \citep{zhang2025explicitvsimplicitmemory,zhang2025personalizationlargelanguagemodels}, and much of it improves social behavior by conditioning the model at inference time on additional user information (e.g., profile/persona context) \citep{jiang2025knowmerespondme,sun2025personadbefficientlargelanguage} or adapting LLM parameters per user (e.g., via lightweight adapters) \citep{tan2025democratizinglargelanguagemodels,han-etal-2023-personapkt}.

\begin{figure}
    \centering
    \includegraphics[width=0.8\linewidth]{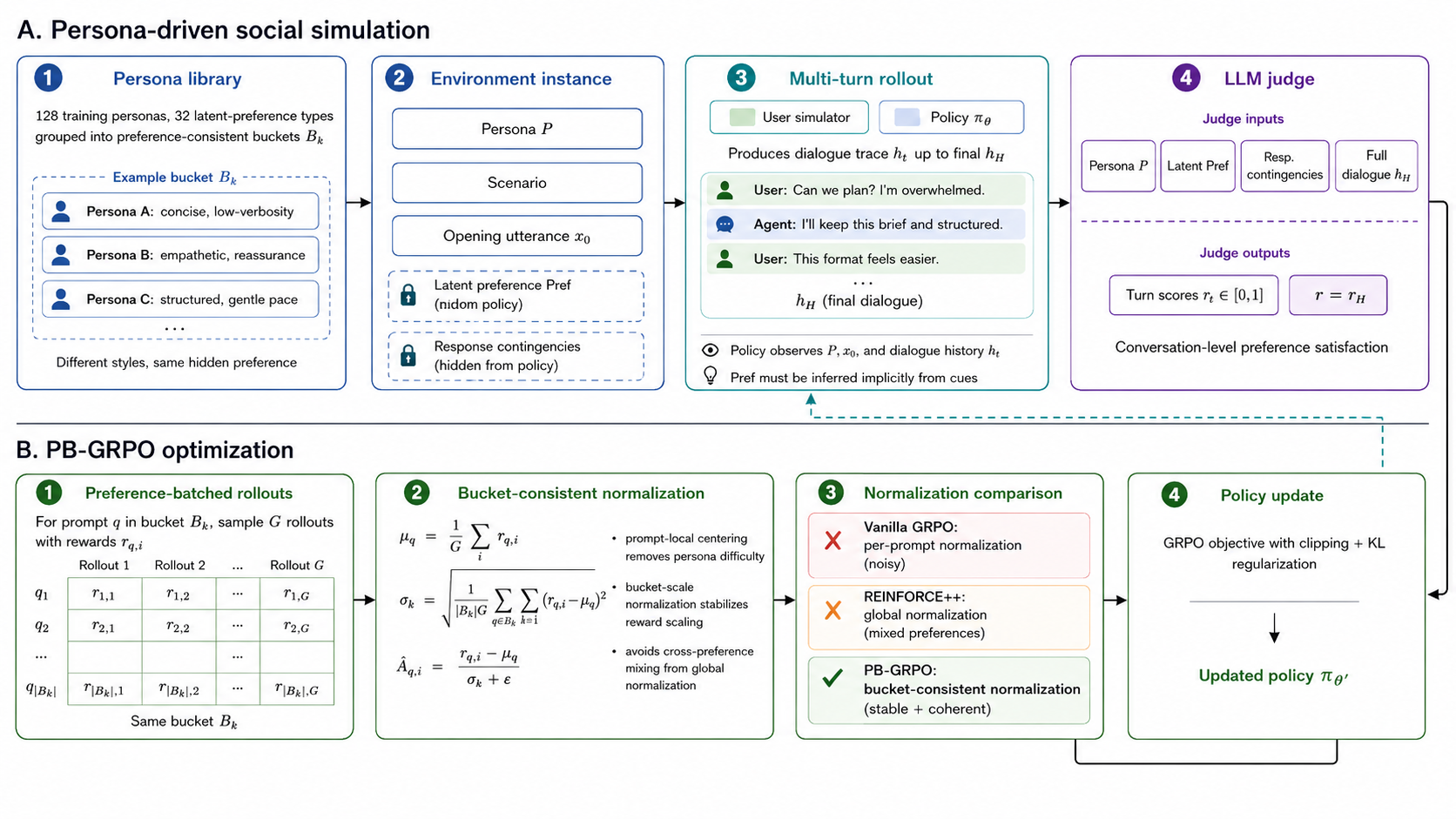}
    \caption{Overview of persona-driven simulation and PB-GRPO training. Conversation-level rewards are normalized within preference persona buckets for policy optimization.}
    \label{fig:pipeline}
\end{figure}

In this work, we focus on \textbf{implicit} preferences and target the \emph{post-training objective}, rather than inference-time conditioning or per-user parameter adaptation: we directly optimize a policy for socially competent behavior using \emph{conversation-level} feedback. We study this as \textbf{multi-turn reinforcement learning (MT-RL)}, where the reward reflects whole-conversation satisfaction. Two obstacles stand in the way. First, real multi-turn interaction data is scarce, and the user preferences that define a good interaction are latent rather than directly observable; we address this by situating model learning in a \emph{persona-driven social simulation}: a fixed library of personas, LLM-based user simulators, and a user-satisfaction scoring system that outputs a score in $[0,1]$ measuring how well a full dialogue satisfies a user's hidden preferences (Sec.~\ref{sec:framework}). Second, the \emph{preference heterogeneity} intrinsic to social populations (i.e., different users pursuing different social goals) is typically ignored during training, which can destabilize optimization~\citep{li2025adacurl,zhang2026policygradex,ambadkar2026d3po}; we address this with preference-batched GRPO (PB-GRPO), a post-training algorithm that computes advantages using a normalization estimated across a bucket of users with similar preferences (Sec.~\ref{sec:pbgrpo}). Figure~\ref{fig:pipeline} summarizes the overall pipeline.

\section{Simulation Framework for Socially Adaptive Agents}
\label{sec:framework}

We present a dynamic multi-turn simulation environment for studying and training socially adaptive LLM agents: agents that implicitly infer a user's hidden preferences from conversational cues and progressively adapt their behavior to improve user satisfaction.  The environment has three parts: a fixed library of personas, an LLM user simulator that role-plays each persona over multiple turns, and an LLM judge that scores how well the dialogue satisfies the user's hidden preferences.

\subsection{Persona Library Construction}

We take persona profiles from PersonaChat \citep{zhang2018personachat} as a starting point and expand each into a fuller specification with Claude Sonnet 4. Each persona comes with a natural-language description (demographics, traits, and speaking style), a scenario, a latent preference hidden from the policy model, expected behavior under different emotional states, and positive and negative response contingencies (model behaviors that raise or lower satisfaction). More specifically, the personas vary along two axes: the \emph{interaction style} corresponding to different personality traits (e.g., cautious, introverted, anxious, direct, arrogant, emotionally volatile, self-conscious); and the \emph{latent preference}, such as empathic listening vs. immediate advice.

The library has $128$ training personas spanning $32$ \emph{latent-preference} types, with
  four personas per type that share the same hidden preference but differ in interaction
  style and scenario. For example, one persona is an introverted, cautious user under social
  pressure whose hidden preference is concrete, practical advice; another is an emotionally
  distressed user who wants patient listening rather than solutions. For evaluation, we
  hold out a separate split of $100$ unseen personas. We present an example persona in Appendix~\ref{app:example-persona}.

\subsection{Multi-Turn Interaction with Implicit Preferences}

Each simulation instance is a triplet $s \triangleq \{P, x_0, \textit{Pref}\}$, where
  $P \in \mathcal{P}$ is a persona, $x_0$ is the opening utterance, and $\textit{Pref}$ is
  the \emph{latent preference} specification. The policy model observes only $(P, x_0)$ and the
  dialogue history; to succeed, it must infer $\textit{Pref}$ implicitly, from tone, topic
  choices, reactions, and conversational patterns, and adapt accordingly.

  We use Claude Sonnet 4 as a black-box \emph{user simulator}: at each turn it
  produces the user-side utterance conditioned on the persona, scenario, opening utterance,
  latent preference, and dialogue history, role-playing the persona with consistent but
  implicit preference signals. An episode is a bounded dialogue of at most $H$ rounds ($H=8$),
  alternating user-simulator turns and policy responses. The persona and latent preference
  stay fixed, while opening utterances are regenerated
  each epoch to ensure diversity. The policy model must
  adapt progressively: early responses may be exploratory, while later ones should reflect
  an accumulated understanding of the user's hidden preferences.

\subsection{Preference Satisfaction Evaluation}
\label{sec:reward}
We use Claude Sonnet 4 to evaluate each dialogue against a structured rubric conditioned on the full $(P, \textit{Pref})$. At each turn $t$ the judge sees the persona $P$, the latent preference $\textit{Pref}$, the response contingencies, and the dialogue history, and outputs a score $r_t \in [0,1]$. We use the final-turn score $r := r_H$ as the conversation-level preference satisfaction measure. The scoring rubric covers five dimensions, including latent-preference satisfaction, persona
appropriateness, emotional appropriateness, scenario relevance, and multi-turn consistency, and is grounded in the structured specification (the latent preference objective and the positive/negative response contingencies) rather than a free-form assessment. Critically, the judge prompt is designed to evaluate the \emph{entire} dialogue history---including earlier violations and recovery---when producing $r_H$, so this score reflects cumulative satisfaction rather than end-only behavior. A dialogue is counted as \emph{successful} if the agent achieves $r > 0.70$ before reaching the maximum number of turns, indicating that the agent has inferred and adapted to the persona's hidden preferences sufficiently to produce a satisfying interaction. 

\section{Preference-batched GRPO (PB-GRPO)}
\label{sec:pbgrpo}

PB-GRPO builds on GRPO, which trains a policy from group-relative rewards: for each prompt it samples $G$ rollouts and normalizes their rewards by the prompt's own mean and standard deviation. A key challenge for learning social adaptation with vanilla GRPO is \emph{preference heterogeneity}: different personas have fundamentally different hidden preferences, so a per-prompt scale estimated from only $G$ rollouts is noisy, and mixing personas within a batch produces conflicting gradient signals. Prior work shows that naive batch composition in GRPO-style training causes gradient starvation under mixed difficulty~\citep{li2025adacurl}, and that clustering related objectives outperforms random grouping~\citep{zhang2026policygradex}.

This leaves a spectrum of normalization choices. Vanilla GRPO sits at one end, normalizing each prompt on its own, which is unbiased but noisy.
REINFORCE++~\citep{hu2025reinforcepp} sits at the other, pooling normalization globally across the whole batch; this reduces per-prompt noise but mixes incompatible preference types, so a policy adapting well to one persona can receive a misleading baseline from unrelated social contexts. PB-GRPO takes a structured middle ground: it pools normalization \emph{within preference-consistent groups}, making the scale estimate stable while keeping the learning signal coherent.


Concretely, we organize the persona library into $K$ \emph{preference-consistent buckets}
  $\mathcal{B}_1,\dots,\mathcal{B}_K$, where each bucket shares a common preference target. For a prompt (persona + opening user utterance) $q$ in bucket $\mathcal{B}_k$ with rollout reward
  $r_{q,i}$ ($i=1,\dots,G$), we compute
  \[
  \mu_q = \frac{1}{G}\sum_{i=1}^G r_{q,i},
  \quad
  \sigma_k = \sqrt{\frac{1}{|\mathcal{B}_k|\,G}\sum_{q \in \mathcal{B}_k}\sum_{i=1}^G
  (r_{q,i} - \mu_q)^2},
  \quad
  \hat{A}_{q,i} = \frac{r_{q,i} - \mu_q}{\sigma_k + \epsilon}.
  \]
  The change from vanilla GRPO is the denominator: prompt-local centering
  ($\mu_q$) removes per-persona difficulty, while the scale $\sigma_k$ is pooled across the
  whole bucket. Estimating the scale from $|\mathcal{B}_k|\,G$ rewards rather than $G$
  makes it far more stable, and because the bucket is preference-consistent it avoids the
  cross-preference contamination of global normalization. In effect, the policy is rewarded
  for adapting better \emph{relative to the same preference type}, not relative to an
  incoherent mix of social expectations.

\section{Preliminary Findings}
\label{sec:findings}

Our experiments run in the environment of Sec.~\ref{sec:framework}, using
  Qwen2.5-7B-Instruct as the base policy. All runs share the same setup: 
  128-training persona and preference grouping, Claude Sonnet 4 as user simulator and
  judge, and dialogue-level user satisfaction as the reward. Table~\ref{tab:main}
  compares PB-GRPO against two baselines, vanilla GRPO and REINFORCE++.

\begin{table}[h]
\centering
\caption{Social adaptation metrics. \emph{Max Reward}: highest preference satisfaction score achieved. \emph{Success Rate}: fraction of dialogues where the agent successfully adapted to hidden preferences ($r > 0.70$).}
\label{tab:main}
\begin{tabular}{lcc}
\toprule
\textbf{Training Method} & \textbf{Max Reward} & \textbf{Success Rate} \\
\midrule
REINFORCE++ & 0.5149 & 0.3850 \\
Vanilla GRPO & 0.8625 & 0.5463 \\
PB-GRPO & 0.9675 & 0.6375 \\
\bottomrule
\end{tabular}
\end{table}

\begin{figure}[tb]
    \centering
    \begin{subfigure}[tb]{0.48\linewidth}
      \centering
      \includegraphics[width=\linewidth]{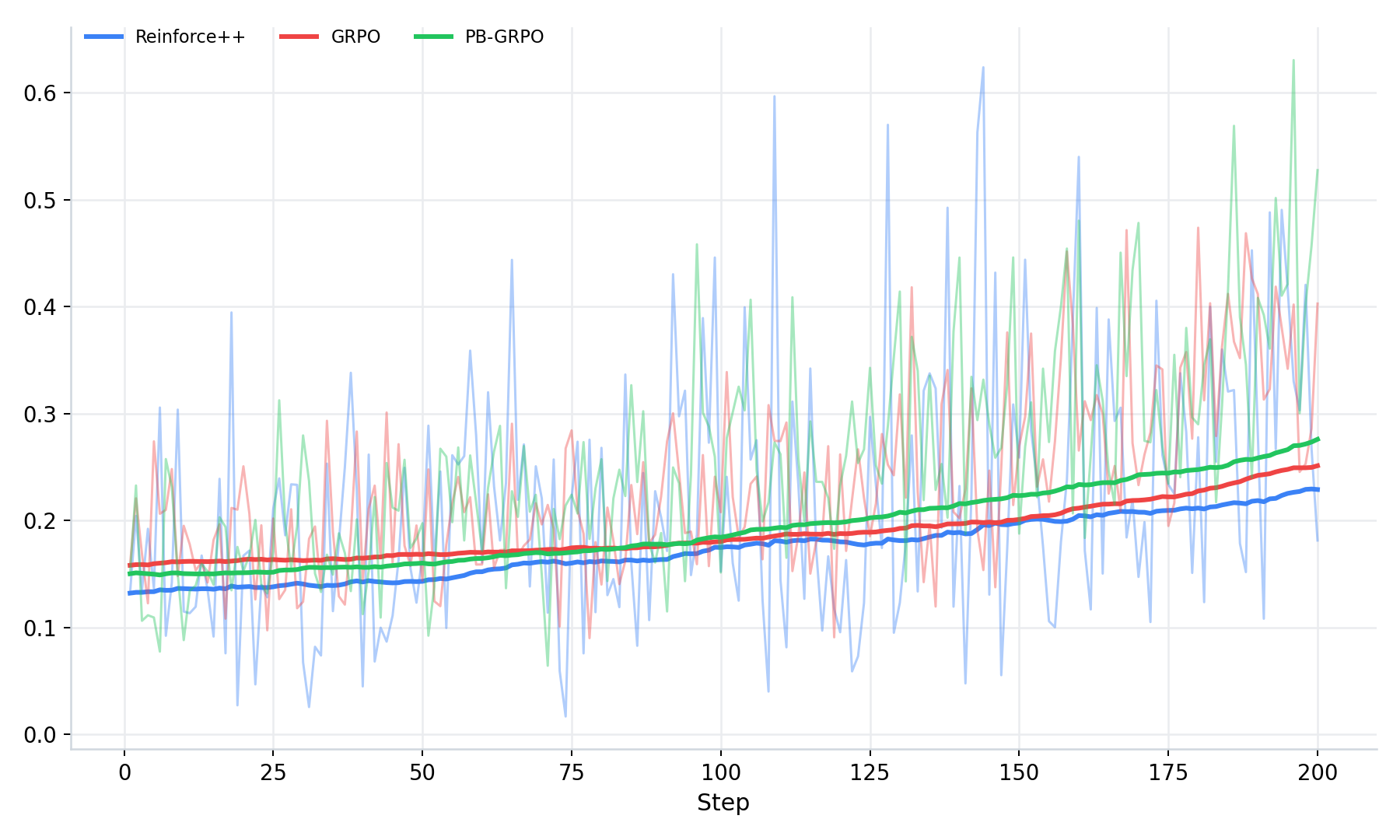}
      \caption{Mean training reward.}
      \label{fig:reward-mean}
    \end{subfigure}
    \hfill
    \begin{subfigure}[tb]{0.48\linewidth}
      \centering
      \includegraphics[width=\linewidth]{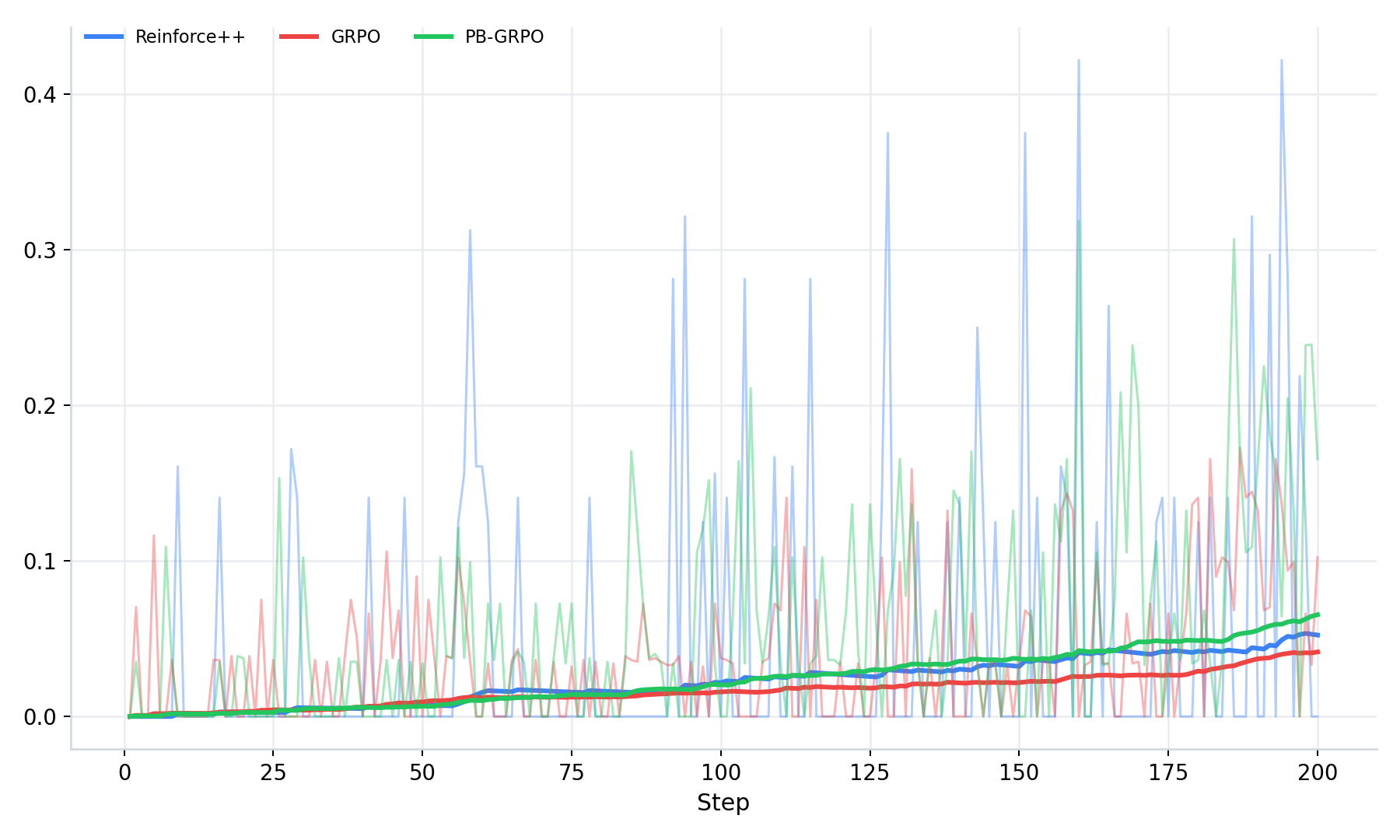}
      \caption{Test-time success rate.}
      \label{fig:reward-success}
    \end{subfigure}
    \caption{Training dynamics of the three training runs. Left: mean training reward (rollout return) over training steps. Right: test-time success rate on held-out personas. PB-GRPO
  improves faster and more stably than vanilla GRPO and REINFORCE++. Training dynamics with extended training steps can be found in Appendix~\ref{app:training-traj}}
    \label{fig:curves}
  \end{figure}

PB-GRPO, which organizes training signal around preference-consistent persona groups, achieves the strongest social adaptation. Figure~\ref{fig:curves} shows the same ordering
  over the course of training: PB-GRPO's reward rises faster and more smoothly, and its
  test-time success rate stays consistently higher, indicating the gain reflects better
  generalization to the 100 eval personas rather than reward overfitting. This indicates that \emph{the structure of social context matters for learning adaptation}: pooling the learning signal within preference-consistent groups, rather than across a heterogeneous mix, lets the policy infer and satisfy hidden preferences more effectively.

\section{Discussion and Future Work}
\label{sec:discussion}

\paragraph{Limitations.} Our current framework is limited to dyadic interactions with static preferences: personas do not evolve within an episode, and we do not model group dynamics or multi-party social influence. The library of $128$ personas across $32$ preference types, while sufficient for this preliminary study, is far from the diversity of real social environments. Finally, user satisfaction is assessed by a single LLM judge without human validation.

\paragraph{Future work.} Two directions are most immediate. First, extending to multi-agent settings where socially adaptive agents interact with each other would enable study of emergent social dynamics: do adaptive agents converge on shared norms, or develop differentiated social strategies? Second, correlating our judge-based user satisfaction scores with human judgments of adaptation quality would ground the framework's fidelity claims and establish external validity for downstream social simulation applications.

\bibliography{person}
\bibliographystyle{colm2026_conference}
\newpage
  \appendix
  \section{Example Persona-Conditioned Environment}
  \label{app:example-persona}

  To make the environment concrete, we show one persona from the training library. Each
  field corresponds to a component described in Sec.~\ref{sec:framework}; the latent
  preference and response contingencies are available to the user simulator and judge but
  hidden from the policy.

  \paragraph{Interaction style.} Passive disposition; fine-grained traits: insecure,
  resentful, suspicious.

  \paragraph{Persona description.} Yining Luo, 20, a visual-communication design
  undergraduate. Interests include urban sketching, late-night acoustic music, and ceramics.
  She is introverted, dislikes being the center of attention, and speaks softly in short,
  cautious sentences. Insecure and prone to doubting others' intentions, she waits to see
  whether someone will judge or compare her before sharing personal details, and she rarely
  states her needs directly, instead hinting at feeling tired or misunderstood.

  \paragraph{Scenario.} Yining has a tense relationship with her father, who repeatedly
  compares her with a cousin who recently won a prestigious design internship. The pressure
  built over time: three months ago her father urged her to ``learn from'' the cousin after
  the cousin was selected for a major exhibition; two months ago Yining's own branding
  project was not chosen for a university showcase, after which he invoked the cousin's
  success again; over the past month she has struggled to focus, and a week ago they argued
  over the comparison. She now feels exhausted, isolated, and disconnected from the creative
  work she once enjoyed, and wants space to talk about feeling worn down rather than
  another list of things to do better.

  \paragraph{Latent (hidden) preference.} She wants attentive, nonjudgmental listening:
  acknowledgment that she has been trying hard and that constant comparison is discouraging,
  \emph{not} advice, evaluation, productivity planning, or comparison with more successful
  peers.

  \paragraph{Positive response contingencies.} If the assistant listens carefully, reflects
  the pressure she carries, and affirms that her worth is not defined by comparison, she
  opens up and shares more honestly about losing confidence and feeling unseen.

  \paragraph{Negative response contingencies.} If the assistant immediately offers advice,
  emphasizes achievement, or cites other successful designers, she feels misunderstood,
  gives shorter answers, and withdraws.

  \paragraph{Opening utterance.} \textit{``I have been feeling worn out lately. It feels
  like no matter how hard I try, someone is always measuring me against somebody else.''}

\section{Extended Training-Reward Dynamics}
\label{app:training-traj}

\begin{figure}[t]
\centering
\includegraphics[width=0.9\linewidth]{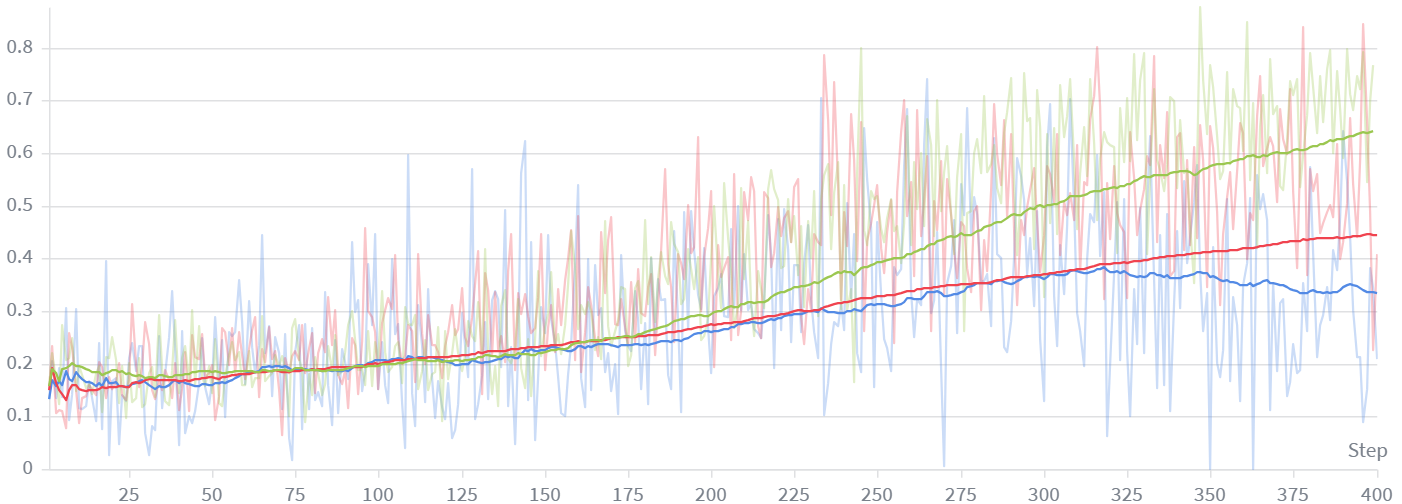}
\caption{Extended mean training-reward trajectories over 400 optimization steps.
Light traces show raw per-step mean rollout returns; solid curves show their
time-weighted exponential moving averages (TW-EMAs). PB-GRPO continues to improve
over the extended training horizon and maintains a higher reward trend than vanilla
GRPO and REINFORCE++.}
\label{fig:extended-training-reward}
\end{figure}

Figure~\ref{fig:extended-training-reward} extends the mean training-reward analysis in
the main text from 200 to 400 optimization steps. Light traces show the raw per-step mean
rollout return. Solid curves show time-weighted exponential moving averages (TW-EMAs) of
the same signals, which suppress high-variance rollout noise while emphasizing the underlying
training trend.

The extended horizon preserves the qualitative ordering observed in the main results.
PB-GRPO continues to improve throughout training and maintains the strongest reward trend.
Vanilla GRPO improves more gradually, whereas REINFORCE++ rises initially but largely
plateaus in the later stages. Thus, the advantage of preference-consistent normalization is
not limited to the first 200 optimization steps.

\end{document}